\pdfoutput=1

\documentclass[11pt]{article}

\usepackage[]{EMNLP2023}

\usepackage{times}
\usepackage{latexsym}
\usepackage{float}

\usepackage[T1]{fontenc}

\usepackage[utf8]{inputenc}

\usepackage{microtype}

\usepackage{inconsolata}
\usepackage{graphicx}
\usepackage[inline]{enumitem}
\usepackage{booktabs}
\title{Radiology-Aware Model-Based Evaluation Metric for Report Generation}

\author{Amos Calamida $^{\clubsuit}$
  \quad {\bf Farhad Nooralahzadeh$^{\clubsuit}$}   \quad {\bf Morteza Rohanian}$^{\clubsuit}$\\
    {\bf Koji Fujimoto}$^{\diamondsuit}$ \quad {\bf Mizuho Nishio}$^{\diamondsuit}$ \quad
  {\bf Michael Krauthammer}$^{\clubsuit}$\\
$^{\clubsuit}$University of Zürich and
University Hospital of Zürich\\
 $^{\diamondsuit}$Kyoto University Graduate School of Medicine\\
 \texttt{amos.calamida@uzh.ch, farhad.nooralahzadeh@uzh.ch}}

\begin{document}
\maketitle
\begin{abstract}
We propose a new automated evaluation metric for machine-generated radiology reports using the successful COMET architecture adapted for the radiology domain. We train and publish four medically-oriented model checkpoints, including one trained on RadGraph, a radiology knowledge graph. Our results show that our metric correlates moderately to high with established metrics such as BERTscore, BLEU, and CheXbert scores. Furthermore, we demonstrate that one of our checkpoints exhibits a high correlation with human judgment, as assessed using the publicly available annotations of six board-certified radiologists, using a set of 200 reports. We also performed our own analysis gathering annotations with two radiologists on a collection of 100 reports. The results indicate the potential effectiveness of our method as a radiology-specific evaluation metric.
The code, data, and model checkpoints to reproduce our findings will be publicly available.
\end{abstract}

\section{Introduction}
Evaluation metrics are essential to assess the performance of Natural Language Generation (NLG) systems. Although traditional metrics are widely used due to their simplicity, they have limitations in their correlation with human judgments, leading to the need for newer evaluation metrics \citep{Blagec2022,Sai2020,Novikova2017}. However, newer metrics have not been widely adopted in the literature due to poor explainability and lack of benchmarking \citep{Leiter2022}. In the medical image report generation domain, several new metrics have been developed, including medical abnormality terminology detection \citep{Li2018}, MeSH accuracy \citep{Huang2019}, medical image report quality index \citep{Zhang2020}, and anatomical relevance score \citep{Alsharid2019}. These metrics aim to establish more relevant evaluation measures than traditional metrics such as BLEU. However, despite their existence, newer publications still rely on traditional metrics, leading to less meaningful evaluations of specialized tasks \citep{Messina2022}.

\begin{figure}[t]
\centering
        \includegraphics[width=7.5cm]{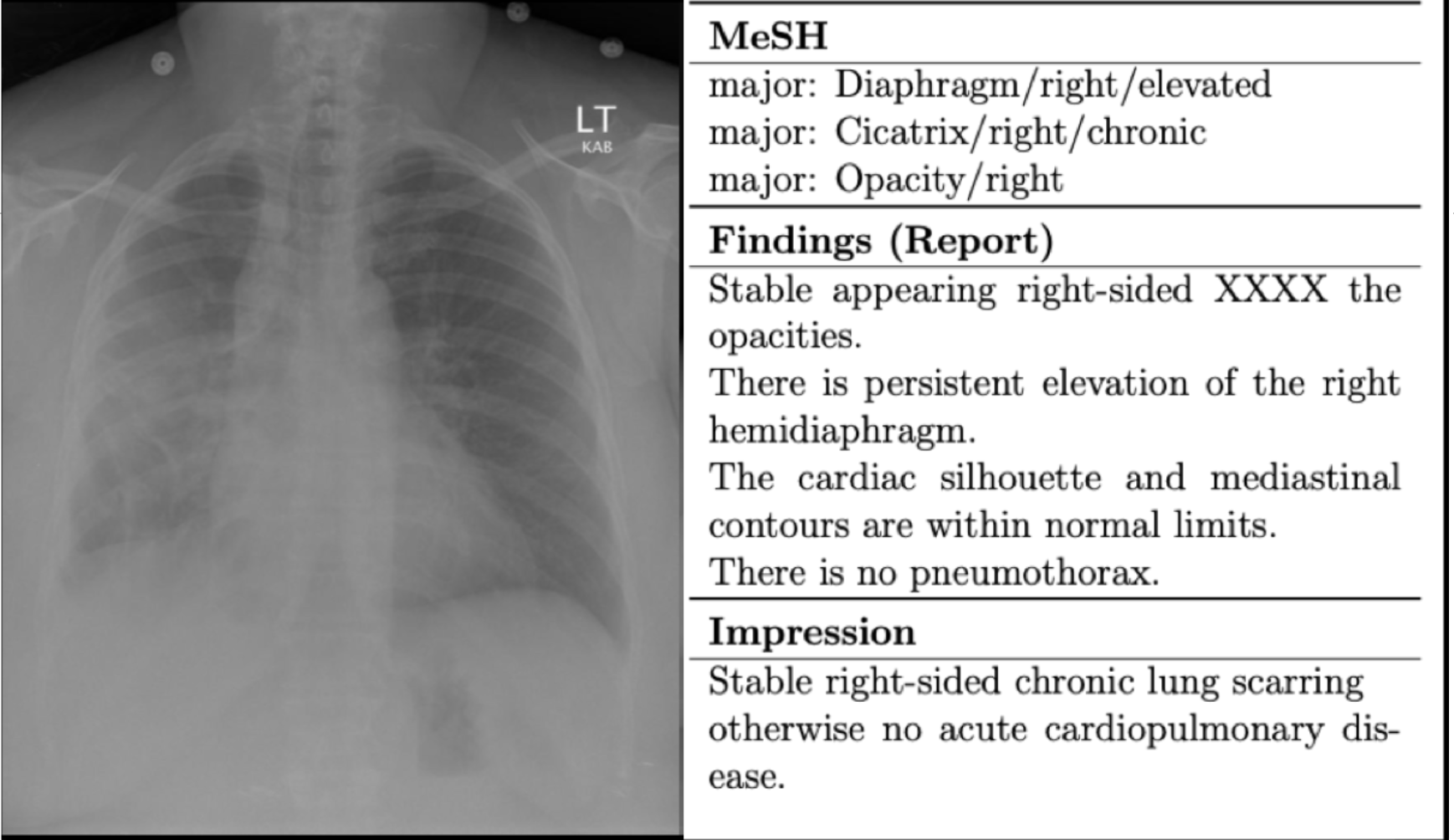}
        \caption{An example report showing the two images and the MeSH, findings and impression columns.\\
        \small Image constructed by the authors with the data from \citet{DemnerFushman2016}.}
    \label{fig:xray_example}
\end{figure}

Radiology reports are narratives that should accurately reflect important properties of the entities depicted in the scan. These reports consist of multiple sentences, including the position and severity of abnormalities and concluding remarks summarizing the most prominent observations (see \autoref{fig:xray_example} for an example report). The task of generating radiological reports is challenging due to their unique characteristics and the need for accurate clinical descriptions \citep{Langlotz2015}. However, current metrics like BLEU do not capture these specific properties, highlighting the need for domain-specific metrics that consider the unique requirements of radiology reports \citep{Chen2020}.
At a high level of abstraction, we attempt to answer the following main research questions in this work:
\begin{enumerate*}[label=(\arabic*), itemjoin={{, }}, itemjoin*={{ and }}]
    \item Can an existing successful metric model architecture be adapted and optimized to develop a novel radiology-specific metric for evaluating the quality and accuracy of automatically generated radiology reports?
    \item To what extent does the integration of radiology-aware knowledge, impact the precision and dependability of the assessment metric in evaluating the efficacy and accuracy of automatically generated radiology reports?
\end{enumerate*}

To this end, we suggest an automated measurement for assessing radiology report generation models. It aims to enhance existing metrics designed for different domains, including both automated metrics like COMET (Crosslingual Optimized Metric for Evaluation of Translation) \citep{Rei2020} and traditional metrics like SPIDEr (Semantic Propositional Image Description Evaluation) \citep{Liu2017} or BLEU \citep{Papineni2002}. This improvement involves incorporating a radiology-specific knowledge graph known as RadGraph \citep{Jain2021}.
Our contributions are as follows:
\begin{itemize}[leftmargin=*]
   \item We design an evaluation model (RadEval) tailored explicitly for assessing radiology reports generated by generative models.
   By incorporating domain-specific knowledge from RadGraph, a radiology-aware knowledge graph, we aim to enhance the accuracy and relevance of the assessment.

    \item We evaluate the proposed strategy by applying it to a set of radiology reports generated by two models. We use the IU X-Ray dataset of ground truth radiology reports and compare the automated scores obtained using our framework with the scores of other established metrics.

    \item We perform an error analysis study with radiology experts that examine the discrepancies between the generated and the ground truth reports. This analysis allows us to further identify the quality of our metric compared with human judgment.

\end{itemize}

\section{Metric Architecture}
\begin{figure}[t]
  \centering
  \includegraphics[width=6cm]{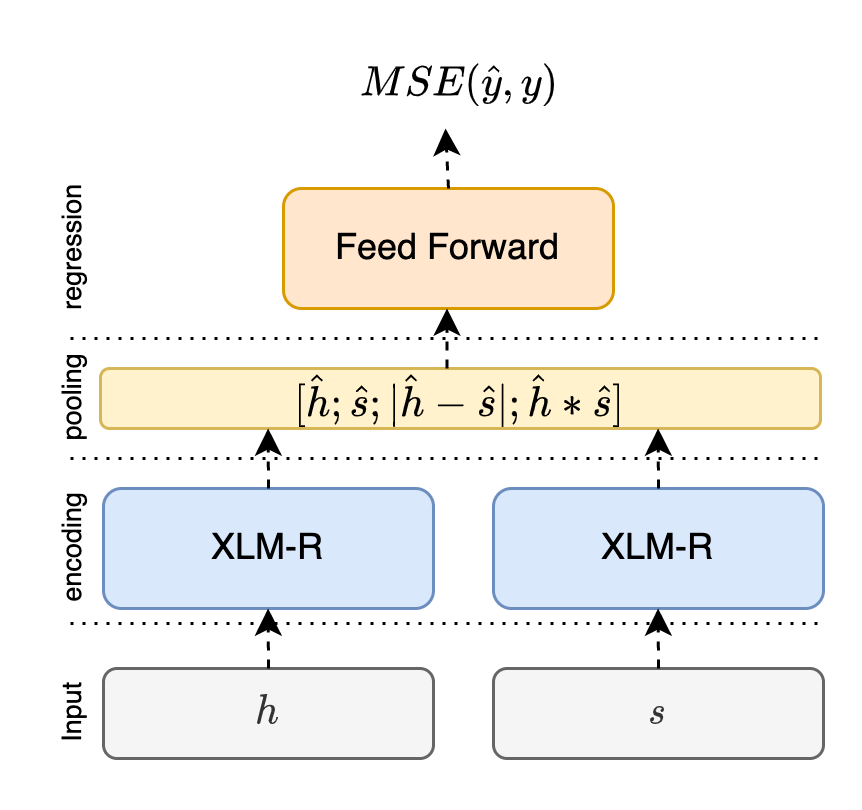}
  \caption{Model architecture for the referenceless metric in the COMET Estimator model (Image provided by \citet{UnbabelCOMET2020}). The source $s$, and hypothesis $h$  are independently encoded using a pretrained language encoder (here it is XLM-R). The resulting embedding vectors are then passed through a pooling layer to create a sentence embedding for each input as $\hat{h}$ and $\hat{s}$.
  Finally, the resulting sentence embeddings are combined and concatenated into one single vector that is passed to a feed-forward regressor. The entire model is trained by minimizing the Mean Squared Error \citep{, Rei2020}.}
  \label{fig:model_architecture}
\end{figure}

We use an evaluation architecture framework developed for machine translation scoring by \citet{UnbabelCOMET2020, Rei2020} and train our own model on radiology data, focusing on the technicalities of radiology reports as outlined before. The framework supports distinct architectures: The Estimator model and the
Translation Ranking model. The fundamental
difference between them is the training objective.
While the Estimator is trained to regress directly on a quality score, the Translation Ranking model is trained to minimize the distance between a “better” hypothesis and both its corresponding reference and its original source.
We use the referenceless mode of the Estimator model as our input data consists of only top inputs - one ground truth report (the source) and one model-generated report (the hypothesis). 
The source $s$, and hypothesis $h$  are independently encoded using a pretrained language encoder (here: XLM-R by \citet{Conneau2020}). The resulting embedding vectors are then passed through a pooling layer to create a sentence embedding for each input.
Given a sentence embedding for the hypothesis  $\hat{h}$, and the source $\hat{s}$, it extracts the following combined features:
\begin{enumerate*}[label=(\roman*), itemjoin={{, }}, itemjoin*={{, and }}]
    \item Element-wise source product: $\hat{h} * \hat{s}$
    \item Absolute element-wise source difference: $|\hat{h} - \hat{s}|.$
\end{enumerate*}
These combined features are then concatenated to the source embedding $\hat{s}$ and hypothesis embedding $\hat{h}$ into a single vector $[\hat{h}; \hat{s}; \hat{h}  *  \hat{s}; |\hat{h} - \hat{s}|]$ that serves as input to a feed-forward regressor. 
Figure~\ref{fig:model_architecture} depicts the COMET Estimator model architecture. The entire model is trained by minimizing the Mean Squared Error (MSE) between the predicted scores and quality assessments (target values) as a loss function.

\section{Dataset Curation} \label{dataset}
Because the COMET architecture is built for assessing the quality of machine translation it requires a parallel corpus of source (i.e. the original text), hypothesis (i.e. the machine translation), and reference (i.e. the correct translation of the source) as input to train the model. In the radiology domain, this corresponds to the source being the ground truth report and the hypothesis being the model generated one. We do not need a reference in our case and are therefore using the referenceless architecture.
To ensure the reliability of our model, we require a sufficiently large number of reports for training. To construct the training data for our metric, we create a corpus of similar reports constructed using the IU X-Ray report collection \citep{DemnerFushman2016}, a widely utilized dataset within the radiology domain. The IU X-Ray dataset contains one to two chest X-Ray scans per data point, along with accompanying reports of actual findings, brief summaries of these findings (referred to as the "impression"), and assigned Medical Subject Headings (MeSH) labels. MeSH is a controlled vocabulary used by the National Library of Medicine database to index and organize biomedical information \citep{NLMMeSH2023}. These terms are used to categorize medical articles based on their content and encompass a broad range of medical topics, including anatomy, diseases, drugs, and procedures.

\begin{figure}[t]
\centering
        \includegraphics[width=7.5cm]{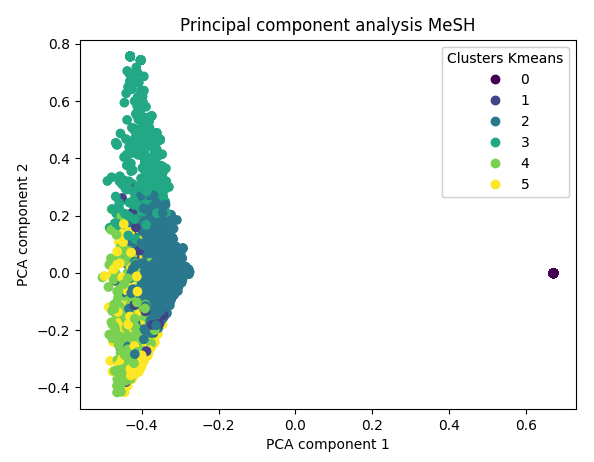}
        \caption{Visualization of the clusters generated by KMeans. The data has been reduced to two dimensions using PCA and the clusters are color-coded. \textbf{MeSH $n=6$}: The outlier (cluster 0) are the normal reports. Clusters 2 and 3 are well defined, 4 and 5 have a lot of overlap. The most prominent values for each cluster can be seen in \autoref{appendix:mesh_cluster_labels}.}
    \label{fig:clusters}
\end{figure}

We concatenated major MeSH labels and removed irrelevant MeSH values (i.e. "no indexing" and "technical quality of image unsatisfactory") for each report. Then we performed a K-Means clustering on the MeSH terms to group reports containing similar topics. To determine the ideal number of clusters we focused on Calinski-Harabasz, as the contents of the MeSH column are more complex and nuanced (i.e. usually anatomy with different quantifiers), and don’t follow a natural sentence structure. We achieved the best results with 6 clusters. This clustering process allowed us to then take the cross-product of each cluster, instead of the entire dataset. \autoref{fig:clusters} shows the final clusters. The most prominent values for each cluster can be seen in \autoref{appendix:mesh_cluster_labels} and the scores to determine the amount of clusters in \autoref{appendix:cluster_scores}.

Next, we scored the similarity of the reports in relation to all other reports in the same cluster using the RadCliQ Metric \citep{Yu2023a}, which is a novel evaluation measure for the similarity of clinical reports leveraging a combination of the BLEU-2 score and the RadGraph F1 metric. The latter ”computes the overlap in clinical entities and relations that RadGraph extracts from a machine- and human-generated reports” \citep[p.4]{Yu2023a}. 

We then generate two sets of comparative report pairs. The first one (referred to as \textit{Best Match corpus} by selecting the top-scored (i.e. most similar) match for each report (based on RadCliQ metric), resulting in a set that encompasses all reports of the cleaned IU X-Ray dataset at least once (i.e. the set size is equal to the size of the cleaned IU X-Ray dataset and each report in the dataset has one corresponding report, which matches best in terms of similarity). The secondary one (referred to as \textit{Top 10\% corpus} allowed for multiple instances of single reports in the set if they had multiple best matches. 

After having created two sets we divided them into two distinct subsets, a training set and a test set. We created a random split of 80/20 using to extract 20\% of the data into the test set and keep the remaining 80\% as the training set. This ensured that our model can be trained and evaluated on two distinct sets of data. With the training process in mind, we also split the training data set further into two subsets, the primary training set, and the validation subset, using the same 80/20 split to have the validation data out of the training set. This validation set is provided to the model trainer to fine-tune its hyper-parameters on each epoch.

During this process, we ensured an appropriate share of normal and abnormal reports are included in both train/validation/test datasets and to not bias the data towards normal reports too much (see \autoref{tab:proportion_normal_abnormal}).
\begin{table}[t]
\begin{tabular}{p{2cm}|cc}
\toprule
\textbf{Dataset} & \textbf{\# reports total} & \textbf{\% abnormal}\\
\midrule
\multicolumn{3}{c}{Top 10\%}\\
\midrule
Training & 47'162 & 63.19\% \\
Testing & 14'738 & 62.19\%\\
Validation & 11'791 & 63.21\%\\ 
\midrule
\multicolumn{3}{c}{Best Match}\\
\midrule
Training         & 471'689                   & 84.64\%              \\
Testing          & 147'403                   & 84.59\%              \\
Validation       & 117'922                   & 84.68\%          \\
\bottomrule
\end{tabular}
\caption{\label{tab:proportion_normal_abnormal}
Size of the different dataset types for the two sets of comparative report pairs and their respective percentage of abnormal reports (i.e. mesh-0 $\neq$ \textit{normal})
}
\end{table}
\section{Model training}

\begin{table*}
\setlength\tabcolsep{3pt}
\centering
\begin{tabular}{p{4cm}|lll|l}
 \toprule
\textbf{Checkpoint Name}	& \textbf{Encoder} & \textbf{Report pairs} & \textbf{Training Target Value} & \textbf{Max($Kendall_\tau$)}\\
\midrule
Match XLM-R RadCliQ & XLM-R & Best Match & RadCliQ-score & 0.696 (Epoch 3)\\
Match Clinic RadCliQ&  BioClinical BERT & Best Match & RadCliQ-score & 0.714 (Epoch 10) \\
\midrule
Top Clinic RadCliQ &  BioClinical BERT & Top 10\%  & RadCliQ-score & 0.830 (Epoch 24)\\
Top Clinic RadGraph&  BioClinical BERT & Top 10\%  & RadGraph F1-score & 0.714 (Epoch 18)\\
 \bottomrule
\end{tabular}
\caption{\label{model-checkpoints}
The specifications of final model checkpoints: {\bf Match XLM-R RadCliQ:} Based on the Best Match corpus, with XLM-R as the encoder layer and RadCliQ as the quality assessments (target values). 
{\bf Match Clinic RadCliQ:} Based on the Best Match corpus, with BioClinical BERT as the encoder layer and RadCliQ as the quality assessments (target values). {\bf Top Clinic RadCliQ:} Based on the Top 10\% corpus, with BioClinical BERT as the encoder layer and RadCliQ as ts the quality assessments (target values). 
{\bf Top Clinic RadGraph:} Based on the Top 10\% corpus, with BioClinical BERT as the encoder layer and RadGraph F1 as the quality assessments (target values). Max ($Kendall_\tau$) is evaluated on the Validation set.
}
\end{table*}
During model training, we optimized the Kendall Tau value between predicted and ground truth rankings. Increasing the maximum number of training epochs from 20 to 40 resulted in higher Kendall Tau values. We also compared the performance of using BioClinical BERT \citep{Alsentzer2019} instead of XLM-R and training on the RadCliQ Score versus the RadGraph F1 score for the Top 10\% corpus.

Our motivation for providing comparative report pairs is to assist future researchers in training their own metrics using a 'Source - Hypothesis' model architecture in their research.
To ensure the quality of our corpus, we have compared the exact overlap on MeSH labels among source and reference reports (i.e. the number of overlapping tokens). Our analysis of the Top 10\% corpus revealed that 80.2\% of the rows had overlap in their MeSH labels, with 46.9\% having one token overlapping and 33.3\% having more than one. Only 19.83\% of rows had no exact overlaps in MeSH tokens. Similarly, when we examined the extent of overlap between MeSH labels in the complete corpus (i.e. among all scores), we found that 34.20\% of rows had no overlap between their MeSH labels. In contrast, 31.69\% of rows had only one overlap, and 34.11\% had more than one overlap between their MeSH labels. We, therefore, see that the scores in the Top 10\% corpus reflect the contents of the reports well. 

\section{Final model checkpoints}
During our experiments with different clustering and similarity score methods, we have generated many comparative report pairs and also already trained several models to benchmark their performance. Out of all models, we have decided to focus on a couple of best-performing checkpoints (based on the highest Kendall $\tau$ value while training).
We used our two corpora (Best Match and Top 10\%) and combined them each once with the XLM-R encoder layer and once with the medical-specific BioClinical BERT \citep{Alsentzer2019}. Also, we trained the models on two scores: Once on the plain Radgraph F1 score, and once on the combined RadCliQ metric score to compare how they differ in correlation performance. 

It is important to notice, that the RadCliQ score is a measure of how many errors a report will contain (i.e. lower is better) and RadGraph F1 is a measure of graph similarity (i.e. higher is better, \citealt{Yu2023a}). Our model checkpoints will behave accordingly when giving their predicted scores. For all checkpoints the scores are unbounded but we provide the typical range.
The names of our checkpoints are based on the type of corpus ({best match} or {Top 10\%}), the encoder ({XLM-R} or {BioClinical BERT}), and the type of score they output ({RadCliQ} or {RadGraph F1}).

We trained the following checkpoints (see also \autoref{model-checkpoints}):
\begin{enumerate*}[label=(\roman*), itemjoin={{, }}, itemjoin*={{, and }}]
\item \textbf{Match XLM-R RadCliQ:} Based on the Best Match corpus, with XLM-R as the encoder layer and RadCliQ as the quality assessments (target values). A lower score indicates a better report. Scores typically fell within -3.5 and +0.5 in our tests. 
\item \textbf{Match Clinic RadCliQ:} Based on the Best Match corpus, with BioClinical BERT as the encoder layer and RadCliQ as the quality assessments (target values). A lower score indicates a better report. Scores typically fell within -3.5 and +0.5 in our tests. 
\item \textbf{Top Clinic RadCliQ:}Based on the Top 10\% corpus, with BioClinical BERT as the encoder layer and RadCliQ as ts the quality assessments (target values).  A lower score indicates a better report. Scores typically fell within -3.0 and +1.5 in our tests. 
\item \textbf{Top Clinic RadGraph:} Based on the Top 10\% corpus, with BioClinical BERT as the encoder layer and RadGraph F1 as the quality assessments (target values). A higher score indicates a better report. Scores typically fell within -0.2 and +1.5 in our test. 
\end{enumerate*}

\begin{table*}
\setlength\tabcolsep{4pt}
\centering
\begin{tabular}{l|cc|c|c|cc}
\toprule
\textbf{Model} & \textbf{BLEU-4} & \textbf{BLEU-2} & \textbf{BERTscore} & \textbf{CheXbert} & \textbf{RadGraph F1} & \textbf{RadCliQ} \\
\midrule
    \multicolumn{7}{c}{Our Top 10\% test data set} \\ 
\midrule
Match XLM-R RadCliQ   & - &\textbf{86.26\%} & 66.98\% & 27.75\% & \textit{71.38\%} & {\color[HTML]{9B9B9B} 95.37\%} \\
Match Clinic RadCliQ   & -  & \textbf{87.99\%} & 67.80\% & 27.80\% & \textit{71.05\%} & {\color[HTML]{9B9B9B} 96.52\%} \\
Top Clinic RadCliQ     & -  & \textbf{88.76\%} & 67.03\% & 27.45\% & \textit{67.22\%} & {\color[HTML]{9B9B9B} 95.51\%} \\
Top Clinic RadGraph    & -  & 41.35\% & \textit{48.86\%} & 24.45\% & {\color[HTML]{9B9B9B} 87.92\%} & \textbf{67.57\%} \\
\midrule
\multicolumn{7}{c}{R2Gen reports} \\ 
\midrule
Match XLM-R RadCliQ   & 78.08\% & \textbf{86.85\%} & \textit{79.54\%} & 52.69\% & 24.74\% & {\color[HTML]{7B7B7B} 66.37\%} \\
Match Clinic RadCliQ     & 81.84\% & \textbf{88.94\%} & \textit{80.95}\% & 51.95\% & 19.36\% & {\color[HTML]{7B7B7B} 63.03\%} \\
Top Clinic RadCliQ       & 77.17\% & \textbf{85.81\%} & \textit{76.63\%} & 47.36\% & 14.52\% & {\color[HTML]{7B7B7B} 58.00\%} \\
Top Clinic RadGraph      & \textbf{61.37\%} & \textit{66.33\%} & 65.09\% & 39.09\% & {\color[HTML]{7B7B7B} 5.17\%} & 40.96\% \\
\midrule
\multicolumn{7}{c}{M2Tr reports } \\ 
\midrule
Match XLM-R RadCliQ  & 74.71\% & \textit{84.88\%} & 76.58\% & 47.66\% & \textbf{85.90\%} & {\color[HTML]{7B7B7B} 95.28\%} \\
Match Clinic RadCliQ    & 79.72\% & \textbf{87.60\%} & \textit{79.83\%} & 45.54\% & 71.73\% & {\color[HTML]{7B7B7B} 87.70\%} \\
Top Clinic RadCliQ      & 73.50\% & \textbf{83.51\%} & \textit{74.55\%} & 43.90\% & 60.46\% &  {\color[HTML]{7B7B7B} 78.58\%} \\
Top Clinic RadGraph     & 58.12\% & \textit{64.29\%} & 64.01\% & 33.64\% & {\color[HTML]{7B7B7B} 65.60\%} & \textbf{71.76\%} \\
 \bottomrule
\end{tabular}
\caption{
Spearman rank correlation between the RadEval score of our model checkpoints and the other metrics based on the generated reports by M2Tr and R2Gen. The \textbf{highest correlation is marked in bold} and \textit{the second highest in italics}. The score on which the specific model checkpoint was trained is printed in {\color[HTML]{7B7B7B}light grey}.
}
\label{model-correlations}
\end{table*}
\section{Model performance}
\label{sec:model_performance}
We assessed the performance of our model's metric using our test dataset (see Section \ref{dataset}) and the IU X-Ray dataset's test set (i.e. 590 sets containing the ground truth and generated reports by two state-of-the-art radiology report generation methods\footnote{We used the following implementations: M2Tr: \url{https://github.com/ysmiura/ifcc} and R2Gen: \url{https://github.com/cuhksz-nlp/R2Gen}}: R2Gen \citep{Chen2020} and M2Tr \citep{Cornia2020}). 
To provide a comprehensive comparison, we calculated the performance of each radiology generation method using five metrics. These involve BLEU \citep{Papineni2002}, BERTScore \citep{Zhang2019a}, CheXbert Similarity \citep{Smit2020}, RadGraph F1 and RadCliQ \citep{Yu2023a}.

BLEU and BERTScore have commonly used metrics in natural language generation tasks to assess the similarity between machine-generated and human-generated texts. BLEU measures the overlap of n-grams and is representative of text overlap-based metrics. On the other hand, BERTScore captures contextual similarity beyond exact textual matches. It uses a pre-trained BERT (Bidirectional Encoder Representations from Transformers) model to encode the two pieces of text and measure their similarity based on their contextualized embeddings. \\
\indent CheXbert vector similarity and RadGraph F1 are metrics specifically designed to evaluate the accuracy of clinical information. CheXbert vector similarity calculates the cosine similarity between the indicator vectors of 14 pathologies extracted from machine-generated and human-generated radiology reports using the CheXbert automatic labeler. This metric focuses on evaluating radiology-specific information but is limited to pathologies. To address this limitation, \citet{Yu2023a} propose the utilization of the report's knowledge graph to represent a wide range of radiology-specific information. Introducing a novel metric called RadGraph F1, they measure the overlap in clinical entities and relations extracted by RadGraph from both machine-generated and human-generated reports. \\
\indent RadCliQ is a combined metric introduced by \citet{Yu2023a}, which combines the BLEU and RadGraph F1 metrics through a linear regression model. The purpose is to estimate the total number of errors that radiologists would assign to a generated report. This metric requires the BLEU and RadGraph F1 scores computed for the generated report as input.
According to \citet{Yu2023a}, RadGraph F1 is the most comparable metric to human judgment, followed by BERTScore, BLEU-2, and CheXbert. We evaluated our model checkpoints trained on RadCliQ and RadGraph F1 to explore their performance difference. We performed the inference using the model checkpoints to obtain the predicted "RadEval" scores. We then calculated the Spearman correlation value between our RadEval Score and the other metrics' scores for the different checkpoints.\\
\indent In the test dataset we constructed (Top 10\%), the data in Table \ref{model-correlations} demonstrates that all RadCliQ-trained models (Match Clinic RadCliQ, Match XLM-R RadCliQ, and Top Clinic RadCliQ) exhibited a high correlation of over 85\% with the BLEU-2 score, which was according to our anticipation as described above. Additionally, these model checkpoints showed the second-highest correlation of approximately 69\% with the RadGraph F1 score, which was also in line with our initial expectations.\\
\indent Interestingly, we found that our RadCliQ-trained models also displayed a reasonably high correlation of approximately 67\% with the BERTscore metric.
The RadGraph F1-trained checkpoint (Top Clinic RadGraph) on the other hand showed the highest correlation with the RadCliQ score at 67.57\% and the second highest correlation with BERTscore at 48.86\%, with BLEU-2 following at 41.35\%. 
It is worth noting that none of our model checkpoints exhibited a high correlation with the CheXbert score, with correlations ranging between 24\% and 28\%. Even though the correlation with BLEU-2 for the RadGraph F1-trained checkpoint was much lower compared to the RadCliQ-trained checkpoints (-45 percentage points), the RadGraph F1-trained checkpoint also showed a lower correlation with BERTscore (-19 percentage points) and CheXbert score (-3 percentage points) at the same time, albeit less drastic than the drop in BLEU-2 correlation.\\
\indent
When we analyzed the correlation scores of our model on the two model-generated datasets. we observed different correlation patterns than the report pairs test dataset. The correlation values between our model and both BLEU scores were high, ranging from 73\% to 86\% on both R2Gen and M2Tr reports for RadCliQ-trained checkpoints. However, for the RadGraph F1-trained checkpoint, the correlation was low, ranging from 14\% to 25\% for R2Gen and 60\% to 85\% for M2Tr. R2Gen had the lowest correlation (5.17\%) with RadGraph F1. The correlation with BERTscore and CheXbert scores was generally higher than the parallel corpus test dataset, ranging from 33\% to 53\% and 64\% to 80\%, respectively.\\
\indent
We found that the RadGraph F1-trained checkpoint for both generation models had better correlation values than the other RadCliQ-trained model checkpoints with BLEU-2 and BERTscore, being at most 19 percentage points away from the highest value for BLEU-2 and at most 11 percentage points for BERTscore. The maximal drop for the CheXbert score was 10 percentage points, compared to 3 percentage points for our corpus test dataset.

\section{Automated metric and radiologist alignment}
In our first experimental alignment study, we make use of the Radiology Report Expert Evaluation (ReXVal) Dataset \citep{Yu2023} \footnote{The dataset became accessible just a day before the submission deadline, leaving us with limited time to utilize it for our evaluation purposes instead of incorporating it into our training process. The Dataset is available on PhysioNet \citep{PhysioNet2000} at \url{https://physionet.org/content/rexval-dataset/1.0.0/}}.
The ReXVal Dataset is a collection of assessments made by radiologists regarding errors found in automatically generated radiology reports. This dataset includes evaluations from six board certified radiologists. The assessments cover clinically significant and clinically insignificant errors, categorized into six different error types. The reports being evaluated are compared to ground-truth reports from the MIMIC-CXR dataset \citep{DBLP:journals/corr/abs-1901-07042}. Each of the 50 studies in the dataset contains one ground-truth report and four reasonably accurate generated reports by selecting candidate reports that score highly according to each of four automated metrics (i.e. BLEU, BERTscore, CheXbert and RadGraph F1), referred to as oracle-metric reports, resulting in 200 pairs of candidate and ground-truth reports that radiologists have annotated.

\begin{figure}[t]
\centering
        \includegraphics[width=.5\textwidth]{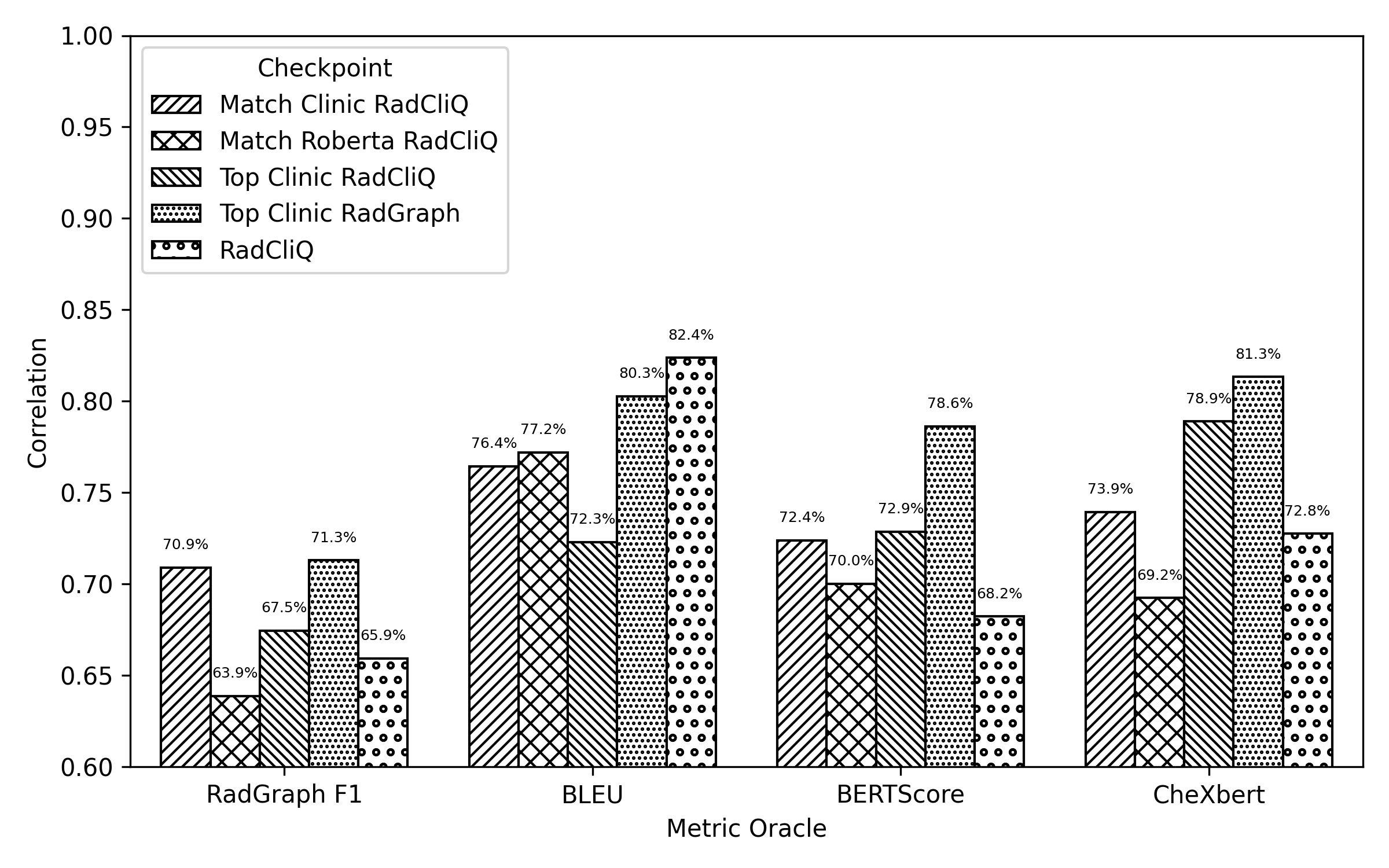}
        \caption{Spearman rank correlation between the RadCliQ \citep{Yu2023a} and our RadEval model checkpoints, and the human error scores assigned by radiologists in four oracle-metric reports datasets}
    \label{fig:correlations_checkpoint}
\end{figure}

We utilized this dataset to assess the correlation between our proposed metric and radiologists' evaluations. To do so,
we employed the approach proposed by the authors to calculate the mean values of significant, insignificant, and total errors for each oracle report, considering the input from their six annotators. Then, we compute RadEval and RadCliQ scores for each metric-oracle report and determine the level of alignment between the radiologists and  the metrics (i.e.RadEval and RadCliQ) using the Spearman rank correlation coefficient.
The results (Figure \ref{fig:correlations_checkpoint}) demonstrate that our proposed metrics, perform better than the compared RadCliQ metric on all oracle reports other than BLEU. Our \textbf{Top Clinic RadGraph} checkpoint is surpassing RadCliQ in terms of correlation with the human up to 10 percentage points (on the BERTscore oracle reports). Also our other checkpoint \textbf{Match Clinic RadCliQ} surpasses the RadCliQ Metric by up to 5 percentage points.


To further investigate the alignment of the automated evaluation metrics with radiologists, we created a balanced dataset of 100 reports for human annotation from an initial set of 590 reports generated using M2Tr \citep{Cornia2020}. The dataset balance was achieved by categorizing the reports into low, average, and high groups based on the 0.33 quantiles of the RadCliQ metric score. Random sampling was then performed to select 150 reports from each category. The reports were further filtered to separate normal and abnormal categories, excluding those labeled as normal in the 'mesh-0' column and removing reports with empty 'mesh-1' values. The remaining abnormal reports were then filtered based on the 'IMPRESSION' column, removing those containing specific phrases associated with normal reports \footnote{i.e., variations of the phrases \textit{no acute cardiopulmonary abnormalities}, \textit{no evidence of active disease}, \textit{no acute findings}. The complete list of filters can be found in \autoref{appendix:filter_phrases}.}. The resulting dataset comprised 80 abnormal and 20 normal reports.

\begin{table}[t]
\scalebox{0.9}{
\begin{tabular}{p{3.1cm}|c|c}
\toprule
\textbf{Model}   & \multicolumn{2}{c}{\textbf{\small Human Annotation Correlation}}\\

& \small \#total errors (\%)      & \small \#sig. errors (\%)\\
\midrule
\multicolumn{3}{c}{\textbf{\scriptsize Complete Human Annotation Dataset: \texttt{100 Examples}}}\\
\midrule
\small BLEU & 17.70   & 13.85\\
\small RadGraph F1  & 28.44    & 16.33            \\
\small RadCliQ        & {\bf 33.49}  & {\bf 19.29}            \\
\midrule
\small Match XLM-R RadCliQ     & 28.71   & 18.37         \\
\midrule
\multicolumn{3}{c}{\textbf{\scriptsize Noisy generation (> 3 errors in the prediction) : \texttt{30 Examples}}}\\
\midrule
\small BLEU         & 27.36   & 
1.39\\ 
\small RadGraph F1         & 19.10   & 0.91           \\
\small RadCliQ         & 33.80   & 1.69           \\
\midrule
\small Match XLM-R RadCliQ    & 34.48    & 6.42  \\    
\small Top Clinic RadCliQ    & {\bf 37.35}   & {\bf 7.97}              \\

\bottomrule
\end{tabular}
}
\caption{\label{model-human-correlations}
Spearman rank correlation between the RadEval score of our two best performing model checkpoints and the human error scores assigned by radiologists. As a comparison, we  include BLEU and two recent radiology-specific metrics and report their correlation scores with our annotators. Correlations for RadGraph F1 are multiplied with $-1$ as these scores estimate the report quality (i.e., higher is better), and the human annotators provide the error score (i.e., lower is better).
}
\end{table}

In this section, we are inspired by the work of \citet{Yu2023a}, where the authors asked a radiologist to count the number of clinically significant and insignificant errors observed in the predicted report for each pair of prediction and ground truth and categorize them into one of the following categories \citep[p.4-5]{Yu2023a} (The categories with $^\dagger$ are added by us).
\begin{enumerate*}[label=(\arabic*), itemjoin={{, }}, itemjoin*={{, and }}]
\item False prediction of finding
\item Omission of finding
\item Incorrect location/position of finding
\item Incorrect severity of the finding
\item Mention of comparison that is not present in the reference impression
\item Omission of comparison describing a change from a previous study
\item $^\dagger$ Mention of uncertainty that is not present in the reference 
\item $^\dagger$ Omission of uncertainty that is present in the reference.
\end{enumerate*} \\
\indent To accomplish the study, initially, two board certified radiologists independently identified and extracted the positive findings from the ground truth reports. The positive findings were then classified into significant and insignificant ones. A comparison was made between the findings extracted from the ground truth reports and the generated reports. Using the eight predefined error categories, the number of errors for each category was counted on the basis of the results of the comparison. Ultimately, both radiologists engaged in discussions with each other and reached a consensus for each report. After receiving the evaluations of two annotators, we evaluated the level of alignment between our metric and their evaluations employing the Spearman rank correlation coefficient. This allows us to quantify the relationship between the metric scores and the count of errors identified by radiologists in the reports. We establish the alignment between the metric and radiologists' evaluations for our checkpoints by conducting this analysis on a selected set of 100 studies. We examine the total number of errors and specifically focus on the number of errors that are clinically significant, as indicated by the radiologists' annotations.
In this analysis, Table \ref{model-human-correlations}, we found that the correlation for the RadCliQ  model is 33.49\% for total errors and 19.29\% for significant mistakes. It shows a slightly higher positive correlation than our two models, indicating a more substantial alignment between the model's predictions and the human-annotated errors in our dataset.
The correlation between Match XLM-R RadCliQ  and human annotation is 28.71\%  for total errors and 18.37\% for significant errors. These values suggest a moderate positive correlation between the model's predictions and the human-annotated total and significant errors. However, it performed up to 11\% better than the compared metrics (BLEU and RadGraph F1) on the total sum of errors and up to 4\% for the significant errors.\\

\indent To further analyze the quality of our metric, we looked at reports with a higher occurrence of errors (referred to as reports with "noisy generation", where our annotators have identified more than 3 errors in total). There are 30 such reports among our 100 studies \footnote{We provided the statistics of the error categories in these 30 examples \autoref{appendix:noisy_statistics} and two noisy examples of what our dataset looks like and the corresponding error categories in \autoref{appendix:noisy_examples}}. When looking only at the noisy reports we can see for the \textbf{Match Roberta RadCliQ} and \textbf{Top Clinic RadCliQ} checkpoints, that we outperform the comparison metrics by up to 19\% on the sum of errors and up to 6\% for the significant errors. For this set of reports, we even perform better than RadCliQ in both categories.

To fairly compare and analyze whether the improvements in the human study (Table \ref{model-human-correlations}; Noisy generation) are statistically significant, we performed the statistically significant test using CoCor \cite{10.1371/journal.pone.0121945} in the ReXVal data set.

Following the CoCor method, we define the following groups in which the groups are dependent and are overlapping: \begin{enumerate*}[label=(\roman*), itemjoin={{, }}, itemjoin*={{, and }}] 
\item JK (Correlation RadCliQ - Human), \item  JH ( Correlation RadEval - Human) \item  KH (Correlation RadEval - RadCliQ)
\end{enumerate*}.
We set \texttt{Alpha = 5\%}, \texttt{Confidence Level = 95\%}, \texttt{Null-Value = 0} and \texttt{Sample size = 50 samples per}.
We have collected the p-values for the dependent, overlapping model according to \citet{hendrickson1970olkin} and report the results for the following set of datasets from ReXVal; where the predictions (generated reports) have been selected based on one of the following Oracle metrics: \begin{enumerate*}[label=(\arabic*), itemjoin={{, }}, itemjoin*={{, and }}]
\item \textbf{BertScore-Oracle}: The null hypothesis can be rejected with p-values 0.0084
\item \textbf{CheXbert-Oracle}: The null hypothesis can be rejected with p-values 0.0131
\end{enumerate*}.
Confirming the alternative hypothesis: r.jk is less than r.jh (one-sided) with r.jk being the correlation of RadCliQ with the human and r.jh being the correlation of RadEval (ours) with the human.

In these two datasets (50 examples each), the correlations are shown in Table \ref{Cocorcorrelations}. The results show that both RadEval and RadCliQ have a strong correlation with human judgments (i.e., > 60\%) rather than our internal data set where both showed moderate correlations (Table \ref{model-human-correlations}).
\begin{table}[t]

\setlength\tabcolsep{4pt}
\scalebox{0.7}{
\begin{tabular}{p{3.3cm}|c|c|c}
\toprule
\textbf{ Dataset name} & \multicolumn{2}{c|}{\textbf{ Correlation}}  & \textbf {p-value} \\
& \small Human-RadCliQ(\%) & \small Human-RadEval(\%) & \\
                      \midrule
{\small ReXVal BertScore-Oracle} & $68.2$&  $78.6$&	$0.0084$\\	
{\small ReXVal CheXbert-Oracle	} & $72.8$&	$81.3$&	$0.0131$
\\	

\bottomrule
\end{tabular}
}
\caption{\label{Cocorcorrelations} Statistically significant test using CoCor \cite{10.1371/journal.pone.0121945} on the ReXVal dataset}

\end{table}
\section{Conclusion}
Our work focuses on developing a novel evaluation metric to evaluate the quality and precision of automatically generated radiology reports. We propose an evaluation model called RadEval that incorporates domain-specific knowledge from a radiology-aware knowledge graph. We train the RadEval model using two corpora, the Best Match corpus and the Top 10\% corpus, which contain pairs of ground truth reports that are similar in terms of their RadGraph representation. We evaluate the performance of the RadEval model on a test set and compare it to other established metrics such as BLEU, BERTScore, CheXbert, RadGraph F1, and RadCliQ. We find that the RadEval model performs well and correlates highly with these metrics. Additionally, when using the new ReXVal dataset of human annotations to compare our alignment with human judgment, we find a high correlation that even surpasses RadCliQ for most report pairs. When conducting our own human annotation study, we did not find a direct high correlation with our human annotators. Still, when comparing with the other metrics' agreement with the same human scores, we also performed better in some cases.
Furthermore, it should be noted that although we have demonstrated relatively strong correlations between automated evaluation metrics and human judgment, additional research is still required to develop an appropriate evaluation metric that aligns with radiologists' expectations and has clinical validity.

\section{Limitations \& Ethical Considerations}
Our proposed method has certain limitations and ethical considerations that merit discussion. 
One limitation of our study is that different radiologists evaluating the reports often gave different scores, even though the effort was to make the evaluation scheme objective and consistent. This variability among radiologists is a common issue when using subjective ratings from clinicians. It suggests that our evaluation scheme may have limitations and it might be challenging to evaluate radiology reports objectively.
Another limitation is that we only considered a specific set of metrics in our study. There are other metrics available that could behave differently than the ones we examined. This means that there could be additional metrics that might provide different insights into evaluating radiology reports.\\
\indent Regarding the datasets used in our study, we exclusively utilized publicly available datasets that are properly anonymized and de-identified, addressing privacy concerns. However, it is crucial to emphasize that if datasets containing comparison exams become available in the future, additional precautions must be taken to ensure that no personally identifiable information is inadvertently disclosed or used in a manner that could identify individual patients The public MIMIC-CXR and IU-X-ray datasets are employed in this work, in which all protected health information was de-identified. De-identification was performed in compliance with Health Insurance Portability and Accountability Act (HIPAA) standards in order to facilitate public access to the datasets. Deletion of protected health information (PHI) from structured data sources (e.g., data fields that provide patient name or date of birth) was straightforward. All necessary patient/participant consent has been obtained, and the appropriate institutional forms have been archived.
We used the datasets for RadGraph and ReXVal, which are under the PhysioNet license. Therefore, as required, we will release our code and data to PhysioNet.\\
\indent By acknowledging these limitations and ethical considerations, we aim to encourage future research and discussions in the field, driving advancements in radiology report generation while prioritizing patient privacy, accuracy, and fairness.




\bibliography{anthology,custom}
\bibliographystyle{acl_natbib}

\clearpage
\appendix

\section{Filter phrases used to get only abnormal reports}
\label{appendix:filter_phrases}

While working on our metric model, we made sure that the input data is balanced in terms of abnormal and normal reports using the following filters. In the first step, we removed all reports with a mesh-0 label of "normal".
In addition, we employed a set of predefined phrases that indicate normal impressions in the radiology reports. These phrases are:
\begin{enumerate*}[label=(\roman*), itemjoin={{, }}, itemjoin*={{, and }}]
\item "No acute cardiopulmonary abnormality"
\item "No acute cardiopulmonary abnormalities"
\item "Negative for acute abnormality"
\item "No evidence of active disease"
\item "No acute cardiopulmonary process"
\item "No acute cardiopulmonary disease"
\item "No acute cardiopulmonary findings"
\item "No acute pulmonary findings"
\item "No acute findings"
\item "No acute cardiopulmonary abnormality identified"
\item "No acute cardiopulmonary abnormality seen"
\item "No acute cardiopulmonary abnormality detected"
\item "No acute cardiopulmonary finding"
\item "No active disease"
\item "No acute disease"
\end{enumerate*}.
\section{Cluster Terms}
\label{appendix:mesh_cluster_labels}
\begin{table}[h!]
    \centering
    \begin{tabular}{c|l}
        \hline
        & \textbf{Most prominent term}  \\
        \hline
        0 & normal                                  \\
        1 & lung/hypoinflation                      \\
        2 & granulomatous disease                   \\
        3 & thoracic vertebrae/degenerative/mild    \\
        4 & calcified granuloma/lung/base/right     \\
        5 & calcified granuloma/lung/upper lobe/left\\
        \hline
        & \textbf{Second most prominent term}\\
        \hline
        0 & No value \\
        1 & lung/hypoinflation markings/bronchovascular \\
        2 & cardiomegaly/mild \\
        3 & thoracic vertebrae/degenerative \\
        4 & calcified granuloma/lung/base/left \\
        5 & calcified granuloma/lung/upper lobe/right \\
    \end{tabular}
        \caption{The most common and second most common terms for each MeSH cluster by numeric cluster Identifier (ID). The principal component analysis of the clusters can be seen in \autoref{fig:clusters}.}

\end{table}

\section{Scores for the MeSH clusters}
\label{appendix:cluster_scores}
\begin{figure}[H]
\centering
        \includegraphics[width=7cm]{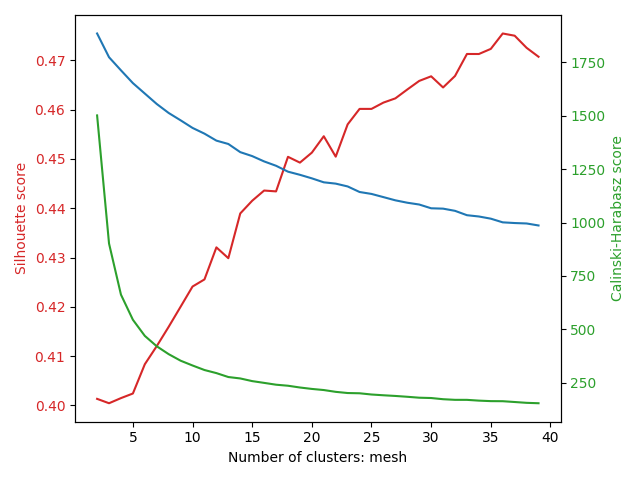}
        \caption{Silhouette score (red), Elbow score/inertia (blue), and Calinski-Harabasz score (green) for increasing amounts of clusters on the MeSH column}
\end{figure}

\section{The statistics of the error categories in the 30 noisy examples}
\label{appendix:noisy_statistics}

\begin{table}[H]
\setlength\tabcolsep{1pt}
\scalebox{0.9}{
\begin{tabular}{p{5cm}|c|c}
\toprule
\textbf{Error Type}  & \multicolumn{2}{c}{\textbf{\small Errors}}\\

& \small \#Significant      & \small \#Insignificant (\%)\\
\midrule
\small(1) False prediction of finding	&10	&7\\

\small(2) Omission of finding	&64	&36\\
\small(3) Incorrect location/position of finding	&1&	1\\
\small(4) Incorrect severity of the finding	&0	&1\\
\small(5) Mention of comparison 	&0	&6\\
\small(6) Omission of comparison	&17	&7\\
\small(7) Mention of uncertainty	&1	&0\\
\small(8) Omission of uncertainty	&3	&0\\
\midrule
Total	&96	&58\\

\bottomrule
\end{tabular}
}
\caption{\label{}
}
\end{table}

\section{Noisy Report Examples}
\label{appendix:noisy_examples}
\paragraph{Example 1.}
\begin{itemize}
\item \textbf{Ground Truth Report}\\\texttt{the heart is enlarged. there is pulmonary vascular congestion with diffusely increased interstitial and mild patchy airspace opacities. the <unk> xxxx pulmonary edema. there is no pneumothorax or large pleural effusion. there are no acute bony findings.}
\item \textbf{Predcition}\\\texttt{there is a right upper lobe opacity. cardiomediastinal silhouette is normal. pulmonary vasculature and xxxx are normal. no pneumothorax or large pleural effusion. osseous structures and soft tissues are normal.}
\end{itemize}

\begin{table}[H]
\setlength\tabcolsep{2pt}
\scalebox{0.8}{
\begin{tabular}{l|c|p{5cm}|c}
\toprule
\textbf{Error Type}  & \textbf{Category}& \textbf {Instances of failure}&\# \textbf {\small Errors}\\
\midrule
\small (2) Omission of Findings & \small Significant&\small	heart size enlarged, vascular congestion, interstital, airspace opacities, pulmonary edema&	$5$\\
\small(3) Incorrect location/position of finding&	\small Insignificant	& \small right upper lobe opacity	&$1$\\
\bottomrule
\end{tabular}
}
\end{table}

\paragraph{Example 2.}
\begin{itemize}
\item \textbf{Ground Truth Report}\\\texttt{stable enlargement of the cardiac stable mediastinal contours. increased interstitial markings in the central lungs and right greater than left. xxxx opacity on the lateral view over the heart also present on the previous exam suggesting chronic subsegmental atelectasis or scarring. no definite pleural effusion seen.}
\item \textbf{Predcition}\\\texttt{the heart and cardiomediastinal silhouette are normal in size and contour. there is no focal air space pleural or pneumothorax. the osseous structures are intact.}
\end{itemize}

\begin{table}[H]
\setlength\tabcolsep{2pt}
\scalebox{0.8}{
\begin{tabular}{l|c|p{5cm}|c}
\toprule
\textbf{Error Type}  & \textbf {Category} & \textbf {Instances of failur}e&\# \textbf \small Errors\\
\midrule
\small (2) Omission of Findings & \small Significant & \small	stable enlargement of the cardiac, stable mediastinal contour, increased interstital markings, xxx opacity - chronic atelectasis or scarring	&$5$\\
\small(6) Omission of comparison&	\small Significant&	\small increased interstitial markings in central lung and right, on the previous exam ...	&$3$\\

\bottomrule
\end{tabular}
}
\end{table}

\end{document}